\documentclass[11pt]{article}
\usepackage{a4wide}
\usepackage{fourier}
\usepackage[utf8]{inputenc}
\usepackage{microtype}
\usepackage{amsmath}
\usepackage{amsthm}
\usepackage{amssymb}
\usepackage{graphicx}

\usepackage{xurl}  % Line breaks urls anywhere

\usepackage[dvipsnames,table,xcdraw]{xcolor}
\usepackage[colorlinks=true,
            citecolor=JungleGreen,
            urlcolor=MidnightBlue,
            breaklinks=true]{hyperref}   
\usepackage[shortlabels]{enumitem}
\usepackage{natbib}

\theoremstyle{plain}
\newtheorem{definition}{Definition}
\newtheorem{theorem}{Theorem}

\newtheorem{proposition}{Proposition}

\newcommand{\vM}{\mathrm{vM}}

\title{Fairness and Randomness in Machine Learning:\\
Statistical Independence and Relativization\footnote{This paper has been presented at
the Philosophy of Science meets Machine Learning Conference in Tübingen in October 2022.}}
\author{Rabanus Derr\\ 
        {\small University of T\"{u}bingen} \\
        \texttt{\scriptsize rabanus.derr@uni-tuebingen.de }\\ 
\and 
      Robert C. Williamson\\
        {\small University of T\"{u}bingen} \\ {\small and T\"{u}bingen AI Centre} \\
        \texttt{\scriptsize bob.williamson@uni-tuebingen.de }
}

\begin{document}

\maketitle

\begin{abstract}
    Fair Machine Learning endeavors to prevent unfairness arising in the context
    of machine learning applications embedded in society. Despite the variety of 
    definitions of fairness and proposed ``fair algorithms,'' there remain unresolved
    conceptual problems regarding fairness.
    In this paper, we dissect the role of statistical independence in
    fairness and randomness notions regularly used in machine learning.
    Thereby, we are led to a suprising hypothesis: randomness
    and fairness can be considered equivalent concepts in machine learning. 
    
    In particular, we obtain a relativized notion of randomness expressed as statistical
    independence by appealing to Von Mises' century-old foundations for probability.
    This notion turns out to be ``orthogonal'' in an abstract sense to the commonly
    used i.i.d.-randomness.
    Using standard fairness notions in machine learning, which are
    defined via statistical independence, we then link the \emph{ex ante}
    randomness assumptions about the data
    to the \emph{ex post} requirements for fair predictions.
    This connection proves fruitful: we use it to argue that 
    randomness and fairness are \emph{essentially} relative and that both
    concepts
    should reflect their nature as modeling assumptions
    in machine learning.
\end{abstract}

\section{Introduction}
Under the name ``Fair Machine Learning'' researchers have attempted to tackle problems of 
injustice, fairness, discrimination arising in the context of machine learning
applications embedded in society \citep{barocas2017fairness}. Despite the variety of 
definitions of fairness and proposed ``fair algorithms,'' we still lack a 
conceptual understanding of fairness in machine learning \citep{passi2019problem, scantamburlo2021non}. What does it mean 
for predictions to be fair? How does the statistical frame influence fairness? 
Is there fair data and what would it look like?

We focus on a collection of widely used fairness notions which are based on 
statistical independence 
e.g.~\citep{Calders2010, Hardt2016, Chouldechova2017}, but examine them from a new 
perspective. Surprisingly, debates concerning these notions
have not questioned the role and meaning of statistical independence upon which they 
are based.
As we shall argue,
statistical independence is far from being a mathematical concept linked
to one unique interpretation (see \S\ref{statistical independence as we know lacks semantics}).
This paper, in contrast to much of the literature on fairness in machine learning, 
e.g.~\citep{Calders2010,Dwork2012, Hardt2016, Chouldechova2017}, investigates what many definitions of fairness
take for granted: a well-defined and meaningful notion of statistical independence.

Another, less popular, strand of research investigates the role of randomness in machine learning \citep{steinwart2009learning}.
The standard randomness notion, independently and often identically distributed data points, suffers the longstanding critique 
of being inadequate (cf. \citep{shafer2007comment}). Again, statistical independence lies at the foundation of this,
hitherto unrelated to fairness, concept. (We will justify below our use of ``randomness'' as
used here.)

At the core of both our observations is the unreflective use
of a convenient mathematical theory of probability. Kolmogorov's axiomatization
of probability theory, developed 1933 in his book
~\citep{kolmogorov1933grundbegriffe}
(translated in~\citep{kolmogorov2018foundations}),
dominates most research in machine learning.
As Kolmogorov explicitly stated,
his theory was designed as a
purely axiomatic, mathematical theory detached from 
meaning and interpretation. In particular, Kolmogorov's
statistical independence lacks such reference.
However, the modeling nature of machine learning
and the arising ethical complications within machine learning applied
in society ask for semantics of probabilistic notions.

In this work, we focus on statistical independence.
We leverage a theory of probability
axiomatized by Von Mises~\citep{mises1964mathematical} in order to obtain
\emph{meaningful} access to probabilistic notions.  (In leaning on Von Mises we are directly following the explicit advice of \citet[page 3, footnote 4]{kolmogorov2018foundations}.)
This theory construes probability theory
as ``scientific''\footnote{We further use the 
term ``scientific'' in order to describe a theory modeling a phenomenon in the world
providing interpretations and verifiability in the sense of Von Mises
(spelled out more concretely in Section~\ref{statistical independence and a probability theory with inherent semantics}). Any more detailed discussion, e.g. along the lines of 
\cite{popper2010logic}, is out of the scope of this paper.} (as opposed to purely mathematical)
with the aim to describe the world and 
provide interpretations and verifiability \citep[pages 1 and 14]{mises1964mathematical}.

Von Mises’ theory of probability provides a mathematical definition of statistical independence
which describes statistical phenomena observable in the world.
In particular, Von Mises' statistical independence is mathematically, but not conceptually,
related to Kolmogorov's definition.

In this paper, we, to the best of our knowledge,
are the first to apply Von Mises' randomness to 
machine learning and to interpret randomness in machine learning in a Von Mises' way.
The paper is structured as follows:

In Section~\ref{a statistical perspective on machine learning}, we outline our
statistical perspective on machine learning. We present
 the ``independent and identically distributed''-assumption (i.i.d.-assumption) as one
commonly used choice for modeling randomness.
The further occurrence of statistical independence
as fundamental ingredient of fairness notions in machine learning
(\S\ref{fairness as statistical indepedencen}) pushes us to the question:
``How to interpret statistical independence in (fair) machine learning?''

We first dissect Kolmogorov's widely used definition
of statistical independence in Section~\ref{independence revisited}
before we propose another mathematical notion following Von Mises.
Von Mises uses his notion of independence in order to define randomness 
% (suprisingly, the concept of randomness is entirely absent from Kolmogorov's theory --- 
% its only indirect appearance is
% in the name ``random variable'', which, in the theory, are neither random nor variables!).
We contrast his definition and the i.i.d.-assumption in Section~\ref{randomness as}.
This reveals a general typification of mathematical definitions of randomness
which most importantly differ in the \emph{absoluteness} respectively \emph{relativity}
to the problem under
consideration (\S\ref{relative randomness instead of absolute, universal randomness}).

Finally, we leverage Von Mises' definition of statistical independence
to redefine three fairness notions from machine learning (\S\ref{von mises fairness}).
Against the background of Von Mises' probability theory, we then link
randomness and fairness both expressed as statistical independence (\S\ref{the ethical implications of modeling assumptions}).
Thereby, we reveal an unexpected hypothesis: randomness
and fairness can be considered equivalent concepts in machine learning.
Randomness becomes a \emph{relative}, even an ethical choice. Fairness, however,
turns out to be a modeling assumption about the data used in the machine 
learning system.

Due to the frequent use of the word ``independence''
with different meanings in this paper, we differentiate.
By ``independence'' we mean an 
abstract concept of unrelatedness and non-influence \citep{OEDindependence}.
We use it interchangeably with ``statistical independence'' which
emphasizes the probabilistic and statistical context.
When referring to the later introduced, formal definitions of
statistical independence
following Kolmogorov or Von Mises we explicitly
state this.
Finally, we assign ``Independence'' (capital ``I'') to one of the fairness criteria in machine
learning which demands for statistical independence of predictions and sensitive attribute
\citep{barocas2017fairness, raz2021group}.
% Generally, this paper can be seen as a contribution to understanding 
% the concept of  ``independence'' itself. 
% Randomness and fairness manifest independence in different context.

%%%%%%%%%%%%%%%%%%%%%%%%%%%%%%%%%%%%%%%%%%%%%%%%%%%%%%%%%%%%%%%%%%%%%%
\section{A Statistical Perspective on Machine Learning}
\label{a statistical perspective on machine learning}
Machine Learning ingests data and provides decisions or inferences. In this sense, 
at its core, it is  statistics\footnote{Machine 
learning could be viewed as classical statistics with 
a stronger focus
on algorithmic realizations 
(cf.~\citep[p. 6]{shalev2014understanding} or \citep{lafferty2006challenges}). While there are ML approaches 
that do not seem statistical in the classical sense (e.g.~worst case online sequence prediction), our general 
description still holds.}.
Adopting this perspective, we understand machine learning
as ``modeling data generative processes''. Statistics, respectively machine learning,
asks for properties of data generating processes given a collection
of data \citep[p. ix]{wasserman2004all} \citep[p. 1]{Bandyopadhyay2011}.

We assume that a data generative process occurs somehow in the world.
We are confronted with a collection of numbers, the data, produced by the process
and acquired by measurement.
A ``model'' is a mathematical description of such a data
generating process. This description should allow us to make predictions 
in an algorithmic fashion. Thus, we require, \emph{inter alia},
a mathematical description of data --- ``a model for data collection''
\citep[p. 207]{casella2002statistical}.

What is arguably \emph{the} standard model of data is stated in \citep[p. 11]{devroye2013probabilistic}
\begin{quote}
    We shall assume in this book that $(X_1, Y_1), ..., (X_n, Y_n)$, the data, 
    is a sequence of independent identically distributed (i.i.d.) random pairs with 
    the same distribution\ldots
\end{quote}
Similar definitions can be found in many machine learning or statistics textbooks
(e.g. \citep[Def. 5.1.1]{casella2002statistical}).
Data (measurements from the world)  is conceived of (mathematically) as a collection of 
random variables which share the same distribution and which are (statistically)
independent to each other. Implicit in this definition is that the data indeed \emph{has} a stable
distribution.
The assumed independence can be interpreted as presumption of
randomness of the data. Each data point was ``drawn independently''\footnote{Observe that as 
    a mathematical theory of data this already leaves a lot to be desired: you will not find in any text
    a precise and constructive explanation of this process of ``drawn independently''. 
    To be sure, the notion
    of ``statistical independence'' is well defined (see further below), as is 
    a collection of random variables. But not the mysterious process
    of ``drawing from''. The closest we can come to such a description of mechanism is the indirect version:
    the data are created (by the world) in a manner that the (mathematical theorem) of the  law of 
    large numbers holds, and that their empirical distribution converges
    to the distribution which was given in the first place.} 
from all others, with the obvious interpretation that each data point does 
not give a hint about the value of any other.

The i.i.d.~assumption is two-fold: 1)  the assumption of identical 
distributions of a sample, and 2) the mutual independence of points in 
the sample. The second assumption alone captures and pertains to
randomness \citep[Section 3]{humphreys1977randomness}. However, since 
the use of i.i.d.~is more
common, we refer to this more specific assumption.

Even though many results in statistics and machine learning rely on the i.i.d. assumption (e.g. 
law of large numbers and central limit theorem in statistics \citep{casella2002statistical},
generalization bounds in statistical learning theory \citep{shalev2014understanding}
and computationally feasible expressions in probabilistic machine
learning \citep{devroye2013probabilistic}),
it has always been subject of fundamental critique\footnote{Nicely summarized by Glenn Shafer in a comment
on \citep{gammerman2007hedging} ``The i.i.d.~case has also been central to statistics ever since Jacob Bernoulli proved the law of large numbers at the end of the 17th century, but its inadequacy was always obvious.'' \citep{shafer2007comment}.}.
Other randomness definitions are rarely
applied, but exceptions exist \citep{vovk1993logic, tadaki2014operational}.

In summary, statistical conclusions often rely on the
i.i.d.-description of data. This description embraces
a model of randomness making randomness a substantial
assumption about the data in statistics and
machine learning.
Interestingly,  statistical independence lies at the foundation of 
another, hitherto unrelated, concept: many fairness criteria in 
fair machine learning are expressed in terms of statistical independence.

%%%%%%%%%%%%%%%%%%%%%%%%%%%%%%%%%%%%%%%%%%%%%%%%%%%%%%%%%%%%%%%%%%%%%%%%%%%%%%%%%%%%%%%%%%%%%%%%%%%%%%%%%%
\section{Fair Machine Learning Relies on Statistical Independence}
\label{fairness as statistical indepedencen}
With the broad use of machine learning algorithms in many socially relevant
domains, e.g. recidivism risk prediction or algorithmic hiring, 
machine learning algorithms turned out to be part of discriminatory practices \citep{Chouldechova2017, raghavan2020mitigating}.
These revelations were accompanied by the rise of an entire research field,
called ``fair machine learning'' (cf. \citep{Calders2010, Chouldechova2017, barocas2017fairness}).
We do not attempt to 
summarize this large literature here. Instead, we simply take a
snapshot of the most widely known fairness criteria in machine learning 
\citep[p.~45]{barocas2017fairness}. 

\subsection{Three Fairness Criteria in Machine Learning}
The three so called observational fairness
criteria, which are expressed in terms of statistical independence, encompass a large
part of fair machine learning literature:
\footnote{Obviously, there are other 
fairness notions in machine learning
which we have not listed above and which are not expressed 
as statistical independence, e.g. \citep{Dwork2012, kilbertus2017avoiding}.}
\begin{description}
    \item[Independence] demands that the predictions $\hat{Y}$
    are statistically independent of group membership $S$ in socially salient, 
    morally relevant groups \citep{Calders2010, Kamishima2012, Dwork2012}.
    \item[Separation] is formalized as $\hat{Y} \perp S | Y$, i.e.~the 
    prediction $\hat{Y}$ is conditionally independent of the
    sensitive attribute $S$ given  the true label $Y$ \citep{Hardt2016}.
    \item[Sufficiency] is fullfilled if and only if
    $Y \perp S | \hat{Y}$ \citep{Kleinberg2017}.
\end{description}
For the sake of distinguishing between the fairness criteria ``Independence'' and
statistical independence, we henceforth mark all fairness criteria by a leading
capital letter.
Each of the notions appear in a variety of ways and under different names \citep[p. 45ff]{barocas2017fairness}.
From the perspective of ethics, the fairness criteria have been substantiated
via loss egalitarianism \citep{binns2018fairness, williamson2019fairness},
 absence of discrimination \citep{binns2018fairness},
affirmative action \citep{Biddle2005, raz2021group} or equality of opportunity
\citep{Hardt2016}.

Certainly, statistical independence is not equivalent to fairness in general 
(a constellation of concepts 
sharing a common name, and perhaps little else agreed by all)
\citep{Dwork2012, Hardt2016, raz2021group}. The nature of fairness
has been discussed for decades, e.g.~in political
philosophy \citep{rawls1971theory}, moral philosophy \citep{lippert2006badness} 
and actuarial science \citep{abraham1985efficiency}. The ``essentially contested''
nature of fairness suggests that no universal, statistical criterion of fairness
exists \citep{gallie1955essentially}. How fairness should be defined is a
context-specific decision \citep{scantamburlo2021non, hertweck2021moral}.

Nevertheless, in order to incorporate fairness notions into algorithmic tools
we require mathematical formalisations of fairness definitions.
The three named criteria dominate
most of the practical fair machine learning tools \citep[p. 45]{barocas2017fairness},
presumably because
their simple definitions make it easy to incorporate them in
learning procedures in a pre-, in- or post-processing way \citep[Chapter 3, p. 20]{barocas2017fairness}.
Regarding both the reductionist definition of fairness
and the pragmatic justification, we emphasize that \emph{our argumentation
is solely with respect to the fairness criteria named above}.

The three fairness criteria are described as
\emph{group} fairness notions since each of the definitions
is intrinsically relativized with respect to sensitive groups.
The definition of sensitive groups substantially
influences the notion of fairness.
For instance, via custom categorization
one can provide fairness by group-design (see 
\citep[Section H.3]{menon2018cost} for a detailed discussion 
of the question of choice of groups).
In addition, the meaning of groups in a societal context
influences the choice of groups
as elaborated in~\citep{hu2020s}, and as explored in a long line of work in social psychology
\citep{Campbell1958, Tajfel1974, HoggAbrams1998, McGarty1999, Castano2002}.
We contribute to 
the debate by drawing a connection between the
choice of groups and the choice of randomness in
\S\ref{randomness as ethical choice}.

%%%%%%%%%%%%%%%%%%%%%%%%%%%%%%%%%%%%%%%
\subsection{Independence in mathematics and the world}
Behind the formalization of
fairness as statistical independence, there is
an apparently rigid, mathematical definition of statistical independence.
The fairness criteria presume that we
have the machinery of probability theory at our disposal and
the relationship of mathematics and the world is clear and unambiguous.
However, as we elaborate further in the following, 
there is no single notion of ``the'' mathematical 
theory of probability \citep{fine2014theories}.
Furthermore, it is not clear what it means to be statistically 
independent when talking of measurements in the world. Respectively, it is not obvious that
the standard formulation of statistical independence
is the right one to use.

If we desire statistical independence to capture a
fairness notion applicable to the world, we ought to
understand what the mathematical formulae signify
in the world. Thus,
in addition to the debate about the fairness criteria,
a debate on the interpretation of statistical concepts
in ethical context is required.

In this work, we contribute to the understanding by 
scrutinizing the standard definition of statistical independence.
Motivated by the occurrence of statistical independence
as fundamental ingredient in randomness as well as fairness
in machine learning, we first detail the standard account
due to Kolmogorov. What is statistical independence? How
does statistical independence relate to an independence
in the world?

%%%%%%%%%%%%%%%%%%%%%%%%%%%%%%%%%%%%%%%%%%%%%%%%%%%%%%%%%%%%%%
\section{Statistical Independence Revisited}
\label{independence revisited}
To make any sense of phenomena in our world, we need to ignore 
large parts of it in order to avoid being overwhelmed. Hence, we
usually assume or presume  the phenomena of interest depends only on a
few factors and to be \emph{independent} of everything else 
\citep{naylor1981decomposition}.
Thus the concept of independence is inherent in a variety of subjects ranging
from causal reasoning \citep{spohn1980stochastic}, to 
logic \citep{gradel2013dependence}, accounting \citep{cook2020concept},
public law \citep{murchison1992concept} and many more.
Independence, as we understand it, grasps the concept of incapability
of an entity to be expressed,
derived or deduced by something else \citep{OEDindependence}.

\subsection{From Independence to Statistical Independence}
Of special interest to 
us is the concept of independence in probability theories and statistics
\citep{levin1980concept}\citep[Section IIF, IIIG and VH]{fine2014theories}.
 Independence in a probabilistic context should somehow capture the unrelatedness between
the occurrence of events, as has been understood for centuries:
\begin{quotation}
    Two Events are independent, when they have no connexion [sic]
    one with the other, and that the happening of one 
    neither forwards nor obstructs the happening of the other.
    \citep[Introduction, p. 6]{DeMoivre1738}.
\end{quotation}
Modern probability theory loosely follows this
intuition as we see in the following.

\subsection{Statistical Independence as we know it lacks semantics}
\label{statistical independence as we know lacks semantics}
Since the axiomatization of probability theory developed by~\cite{kolmogorov1933grundbegriffe}
(translated in~\citep{kolmogorov2018foundations}),
it displaced many other approaches.
Mathematically, Kolmogorov's measure-theoretic
axiomatization 
dominates all other mathematical formalizations to probability and related
concepts.\footnote{Exceptions exist,
e.g. in quantum theory \citep{gudder1979stochastic} or in statistics \citep{vovk1993logic}.}
In particular, his definition of statistical independence
developed a well-accepted, ubiquitous notion.
In a simple form it is given by\footnote{For a rigorous definition see 
Appendix~\ref{def:kolomogorov independence}.}:
\begin{definition}[Simplified Statistical Independence following Kolmogorov]
\label{def:simple-stat-independence-kolmogorov}
    Two events $A, B$ are statistically independent iff
    \begin{align*}
        P(A\cap B) = P(A) P(B).
    \end{align*}
\end{definition}
Independence plays a central role in Kolmogorov's probability theory:
``measure theory ends and
probability begins with the definition of independence'' 
\citep[p. 37]{Durrett:2019tt} (quoted in \citep{Tao:2015ur}).
However, Kolmogorov's
definition of independence is subtle and requires closer investigation.

Let us look onto a small toy example: consider the
experiment of throwing a die. The events under observation 
are $A = \{ 1,2\}$, seeing one or two pips, respectively
$B = \{2,3\}$, seeing two or three pips.
If the die were fair, so each face has equal probability $\frac{1}{6}$
to show up, the events $A$ and $B$ would turn out not to
be independent. In contrast, if the die were loaded in a very special way
$p_1 = \frac{1}{3}, p_2= \frac{1}{6}, p_3=\frac{1}{2}, p_4=p_5=p_6=0$, where
$p_i$ refers to the probability of seeing $i$ pips, the events $A$ and $B$
would be independent following Kolmogorov's definition.

Thus, statistical independence, even though defined over events,
manifests in the correspondence of how probabilities
are mapped to events and why. The definition
focuses on events. But, the crucial ingredient is 
the probabilities. Thus, there is no unique interpretation
of statistical independence. A more detailed interpretation
and meaning heavily depends on the
interpretation of probability in the first place.

Given this observation, we may ask for a notion of independence
in the world, which is somehow captured by
Kolmogorov's definition. Kolmogorov
himself underlined the avowedly mathematical, axiomatic
nature of probability. His theory is in principle
detached from any meaning of probabilistic concepts such
as statistical independence
in the world \citep[p. 1]{kolmogorov2018foundations}.
He even questions the validity of his axioms as reasonable descriptions of
the world \citep[p. 17]{kolmogorov2018foundations}.

However, one can possibly construct a notion of  independence
in world which is captured by Kolmogorov's
definition. If one assumes one's calculations about one's beliefs
on the happening of events
are governed by the mathematical rules laid out by Kolmogorov, then
Kolmogorov's definition of statistical independence
captures one's (in-the-world) understanding of an independence
of beliefs on the happening of events.
This sketch of a purely subjectivist account to
statistical independence neglects a justification
for the choice of mathematical formulation and
skips over any reference to an objective world.
In conclusion, Kolmorogov's independence might capture a worldly concept.
But, this independence in the world is not 
uniquely attached to Kolmogorov's definition.

There is a third major irritation arising from
Kolmogorov's definition.
As observed already above, Kolmogorov
treats events as the entities of independence.
Though, against the background of \citet{DeMoivre1738}'s 
intuition on statistical independence, we wonder about this focus.
Statistical independence, as \citet{DeMoivre1738} already emphasized,
refers to altering ``the happening of the event,'' but not the event itself.
It is not the independence between the shown numbers of the die (whatever
this means), but the independence of the processes how the number
showed up (loading the die, throwing the die, etc.) which are
captured by statistical independence. 

This critique is not new. Von Mises already criticized
the measure theoretical definition by Kolmogorov in his
book \citep[p. 36-39]{mises1964mathematical}. In summary,
he argued that there is
no interpretation of statistical
independence of ``single events.'' The unrelatedness, which the probabilistic
notion of statistical independence is trying to capture, locates in the process of
reoccuring events, but not single events themselves.
More recently, \citet{Collani2006} argued in a similar way.
It is probability which brings the definition of statistical independence to life
and it is the question what probability means and why we use it
which links the mathematical definition to a concept in the real world.

In machine learning and statistics it is often presumed that 
the mathematical definition of 
independence captures a worldly concept.
As we argued in this section, this link is far from being well-defined.
However, if we consider machine learning as worldly data modeling, then
the natural question arises: what do we model when we leverage Kolmogorov's statistical independence?
What do we mean by independence of events?
In order to circumvent these questions, we propose to look into another
mathematical theory of probability.
This theory was led by the idea of modeling
statistical phenomena in the world.

%%%%%%%%%%%%%%%%%%%%%%%%%%%%%%%%%%%%%%%%%%%%%%%%%%%%%%%%%%%%%%%%%%%%%%%%
\subsection{Statistical Independence and a Probability Theory with inherent Semantics}
\label{statistical independence and a probability theory with inherent semantics}
Around 15 years before Kolmogorov's 
\emph{Grundbegriffe der Wahrscheinlichkeitsrechnung}
\citep{kolmogorov1933grundbegriffe} 
(translated as \citep{kolmogorov2018foundations}),
Von Mises proposed an earlier
axiomatization of probability theory \citep{mises1919grundlagen}. 
His less known theory approached the problem of a mathematical theory 
of probability through the lens of physics.
Von Mises aimed for a ``mathematical theory of repetitive events.''

This aim included the emphasis on the link between real-world
and mathematical idealization. In particular, he offers interpretability and
verifiability of his theory
\citep[p. 1 and 14]{mises1964mathematical}.
\footnote{
Von Mises' discussion pre-dates most of the modern work done in the philosophy of science.
\citet{popper2010logic},
for instance, was inspired by Von Mises' conception of randomness and probability.
While  Von Mises' definition of verifiability and interpretability is arguably somewhat ill-conceived,
nevertheless, he works, in contrast to Kolmorogov's theory
which is of purely syntactical nature,
on a semantical project, where the connection of the real world
and the mathematical formulation
is of vital importance.
}
For interpretation he defined
probabilities in a
frequency-based way (see Definition~\ref{def:collective}). This inherently reflects
the repetitive nature of the
phenomena under description.
By verifiability he referred to the ability to \emph{approximately} 
verify the probabilistic statements made about the world \citep[p. 45]{mises1964mathematical}. 

In summary, Von Mises' theory, in our conception of machine learning,
starts the ``modeling of data generating processes'' on an even more fundamental
level then it is currently done via the use of Kolmogorov's axiomatization.
His aim for a mathematical description of data-generating processes
(the sequence of repetitive events) aligns to our perspective on
machine learning as laid out earlier (cf. Section~\ref{a statistical perspective on machine learning}).
With Von Mises we obtain access to \emph{meaningful} foundations
for statistical concepts in machine learning. In particular, we redefine and 
reinterpret
statistical independence in a Von Misesean way. This suggests new perspectives
on the problem of fair
machine learning and the concepts of fairness and randomness themselves.

For the further discussion, we summarize the major ingredients of Von
Mises' theory. Fortunately, it turns out that Von Mises' notion of 
statistical independence, central to our discussion, is 
mathematically analogous to the well-known Kolmogorovian
definition. Thus, one's intuition on statistical independence
is refined but its mathematical applicability remains.

\subsection{Von Mises' theory of Probability and Randomness in a Nutshell}
\label{Von Mises theory of probability and randomness in a nutshell}
Von Mises' axiomatization of probability theory is based on random 
sequences of events and the interpretation
of probability as the limiting frequency that an event occurs in such
a sequence \citep{mises1919grundlagen}.
These random sequences, called collectives, are the main ingredients of
his theory. For the sake of simplicity, we stick
to binary collectives. Thus, collectives are 0-1-sequences with
certain randomness properties which
define probabilities for the labels 0 and 1.
Nevertheless, it is possible to define
collectives respectively probabilities on richer label sets. Collectives can be extended
up to a
continuum \citep[II.B ]{mises1964mathematical}.

For notational economy, we note that a  sequence taking values in $\{ 0,1\}$,  
$(s_i)_{i\in\mathbb{N}}$ can be identified with
a \emph{function} $s\colon\mathbb{N}\rightarrow \{ 0,1\}$.
\begin{definition}[Collective \protect{\citep[p. 12]{mises1964mathematical}}]
\label{def:collective}
Let $\mathcal{S}$ be a set of
sequences $s\colon \mathbb{N} \rightarrow \{ 0,1\}$ with $s(j) =
1$ for infinitely many $j \in \mathbb{N}$. In mathematical terms,
collectives with respect to $\mathcal{S}$ are sequences $x \colon \mathbb{N} \rightarrow \{ 0,1\}$
for which the following two	conditions hold.
\begin{enumerate}
    \item \label{def:probablity a l Von Mises} The limit of relative frequencies of~$1$s,
\begin{align*}
    \lim_{n \rightarrow \infty} \frac{|\{i\in\mathbb{N}\colon  x(i) = 1, 1 \le i \le n\}|}{n}
\end{align*}
exists.
\footnote{The mathematically inclined reader might notice the requirement for an
order structure to obtain a definition of limit here. We use $x$ as sequences
with the standard order structure on the natural numbers. However, $x$ can be
generalized to be a net on more abitrary base sets \citep{ivanenko2010decision} .
}If it exists, then the limit of relative frequencies
of~$0$s exists, too. We define $p_1$, respectively $p_0 = 1 - p_1$, to be its value. 

\item \label{def:randomness a la Von Mises}For all $s \in \mathcal{S}$,
\begin{align*}
    \lim_{n \rightarrow \infty} \frac{|\{i\in\mathbb{N}\colon   x(i) = 1 \text{ and } s(i) = 1,
    1 \le i \le n\}|}{|\{j\in\mathbb{N}\colon  s(j) = 1, 1 \le j \le n\}|} = \lim_{n
    \rightarrow \infty} \frac{|\{i\in\mathbb{N}\colon  x(i) = 1, 1 \le i \le n\}|}{n} = p_1.
\end{align*}

\end{enumerate}
\end{definition}
We call $p_0$ the probability of label $0$. Conversely, $p_1$
is the probability of label $1$.\footnote{Von Mises called $p_0$ chance as long as $x$ is
a sequence. When $x$ is a collective, he called it probability.}
 The existence of the limit (Condition~\ref{def:probablity a l Von Mises})
 is a non-vacuous condition. One can easily
 construct sequences whose frequencies do not converge.

The sequences $s \in \mathcal{S}$ are called selection rules. A selection
rule selects the $j$th element of $x$ whenever $s(j)
= 1$.\footnote{\label{footnote:vonmises-selectionrules}
We sweep under the carpet a substantial difference in Von Mises' definition
of selection rules and our definition. Von Mises allowed the selection rules
to ``see'' the first $n$ elements of the collective when deciding whether to
choose the $n+1$th element \citep[p. 9]{mises1964mathematical}. 
Our definition is more restrictive. We require
the selected position to be determined before
``seeing'' the \emph{entire} sequence.
We focus on the ex ante nature of Von Mises randomness but neglect
his recursive formalism.
}
Informally, a collective (w.r.t~$\mathcal{S}$) is a sequence which has invariant frequency limits
with respect to all selection rules in $\mathcal{S}$. 
We call any selection
rule which does not change the frequency limit of a collective
\emph{admissible}. This invariance property of collectives is often
called ``law of excluded gambling strategy'' \citep{mises1964mathematical}.
When thinking of a sequence of
coin tosses, a gambler is not able to gain an advantage by just considering specific
selected coin tosses. The probability of seeing ``heads'' or ``tails'' remains
unchanged.

Von Mises introduced the ``law of excluded gambling strategy''
with the goal to define randomness of a collective
\citep[p. 8]{mises1964mathematical}. A collective
is called \emph{random} with respect to $\mathcal{S}$. 
Consequently, Von Mises integrated randomness and
probability into one theory.
In fact, admissibility of selection rules is equivalent to statistical independence
in the sense of Von Mises. But it is defined with respect to collectives instead of selection rules.
\begin{definition}[Von Mises' Definition of Statistical Independence of Collectives  \protect{\citep[p. 30 Def. 2]{mises1964mathematical}}]
\label{def:independence Von Mises}
A collective $x$ with respect to $\mathcal{S}_x$ is called statistically independent
to the collective $y$ with respect to $\mathcal{S}_y$
 iff the following limits exist and
\begin{align*}
    \lim_{n \rightarrow \infty} \frac{|\{i \in\mathbb{N}\colon  x(i) = 1 \text{ and } y(i) = 1,
    1 \le i \le n\}|}{|\{j\in\mathbb{N}\colon  y(j) = 1, 1 \le j \le n\}|} = \lim_{n
    \rightarrow \infty} \frac{|\{i\in\mathbb{N}\colon   x(i) = 1, 1 \le i \le n\}|}{n} = p_1,
\end{align*}
where $\frac{0}{0} := 0$. When two collectives are independent of each
other we write 
\begin{align*}
    x \perp y.
\end{align*}
\end{definition}
In comparison to admissibility, the collective $y$
adopts the role of a selection rule. It \emph{is} in fact an admissible
selection rule with the difference that a potentially finite number of
elements in  $x$ are selected (cf. \citep[p. 120f]{gillies2012objective}).
Conversely, Von Mises' randomness is statistical independence with respect
to sequences with
infinitely many ones and potentially no frequency limit.
 (For a general comparison between Kolmogorov's and Von Mises' theory of probability
 see Appendix~\ref{generalized Von Misesean probability theory} and 
Table~\ref{tab:kolmogorov-vs-vonmises-summary}.)

\subsection{Kolmogorov's Independence versus Von Mises' Independence}
\label{Kolmogorov's Independence versus Von Mises' Independence}
What is the relationship between Kolmorogov's and Von Mises'
definition of statistical independence?
On a conceptual level, the critique posed earlier, which questioned the meaning
of statistical independence between events following Kolmogorov, gets
resolved.

Von Mises adopted a strong frequential perspective on probabilities
which clarifies the mapping from real world to mathematical definition.
He idealized repetitive observations by infinite sequences and defined
probabilities as limiting frequencies.\footnote{
Without the idealization, again the mathematical description
would miss a link to a worldly phenomenon. The idealization in
terms of infinite sequences is substantial. In fact,
the legitimacy of this idealization is the subject of 
another debate \citep{hajek2009fifteen}.
Nevertheless, the idealization taken in Von Mises' 
framework is explicitly and transparently stated. Kolmogorov's axioms
do not possess such a statement.}
Von Mises' independence states that there is no difference
in counting the frequency of occurrences of an event in the entirety of the sequences
or in a subselected sequence. His independence forbids any
statistical interference between
processes described as sequences. No statistical patterns can be derived from one
sequence by leveraging the other. Von Mises' definition formalizes the
concept of statistical independence
between processes of reoccuring events.

In contrast to Kolmogorov, Von Mises' definition does not evoke
the conceptual obscurity. His focus on idealized sequences
of repetitive events restricts his definition of statistical independence
to specific applications with the gain
of clarity in the goal of the mathematical description.

On a more formal level, Kolmogorov defined statistical independence
via the factorization of measure
(cf.~Definition~\ref{def:kolomogorov independence}), whereas Von Mises
defined statistical independence via conditionalization of measures. The invariance of the
frequency limit of a collective with regard to the subselection
via another collective can be interpreted as the invariance of a probability of an event
with regard to the conditioning on another event, i.e. ``selecting
with respect to'' is ``conditioning on'' (cf.~Theorem~\ref{thm:admissability gives stat independence}
and Theorem~\ref{thm:independent events give admissible selection rules}).

Mathematically, it turns out that Kolmogorov's definition and Von Mises'
definition are both special cases (modulo the measure zero problem in conditioning)
of a more general form of measure-theoretic statistical
independence. A selection rule with converging frequency limit is
admissible (respectively, statistically independent), to a collective if
and only if the two are statistically independent of each other in the sense of
Kolmogorov, when generalized to finitely additive probability spaces 
(see Appendix~\ref{generalized Von Misesean probability theory} for 
a formal statement of this claim). 
Thus, we can replace the known definition of statistical independence
by Kolmogorov with the definition by Von Mises. Thereby, we give a specific
meaning to statistical independence.

We have been motivated to dissect the notion of statistical
independence for its central role in fair machine learning.
Von Mises' definition drew us closer to a more transparent
mathematical formalization of  statistical independence
for fairness notions in machine learning.
However, our discussion of Von Mises' theory
skipped over a substantial part of his work so far.
Von Mises included a definition of randomness in his theory
of probability. This is much in contrast to Kolmogorov:
There is no definition 
of ``randomness'' in Kolmogorov's
\emph{Grundbegriffe der Wahrscheinlichkeitsrechnung}
\citep{kolmogorov1933grundbegriffe} 
(translated as \citep{kolmogorov2018foundations}).
Even more interestingly, Von Mises' definition 
of randomness is stated in terms of statistical independence.
The reader might notice that in Section~\ref{a statistical perspective on machine learning}
we already stumbled upon a heavily used notion of randomness
in machine learning, which is expressed as statistical independence (i.i.d.).
How do {\frenchspacing i.i.d. and} Von Mises' randomness relate to each other?
How does the close connection
between statistical independence and randomness
complement our picture of the three fairness criteria
from machine learning?

%%%%%%%%%%%%%%%%%%%%%%%%%%%%%%%%%%%%%%%%%%%%%%%%%%%%%%%%%%%%%%%%%%%%%%%%%%%%%%%%%%%%%%%%%%%%%%%
\section{Randomness as Statistical Independence}
\label{randomness as}
The nature and definition of randomness seems as ``random'' as
the term itself \citep{sep-chance-randomness, ornstein1989ergodic, muchnik1998mathematical, volchan2002random,berkovitz2006ergodic}.
Usually, a very broad distinction between two approaches to randomness
is made: process randomness versus outcome randomness \citep{sep-chance-randomness}.
In this work, we focus on outcome randomness
and more specifically the role of randomness in statistics and machine learning.

Randomness is a modeling assumption in statistics
(cf. Section~\ref{a statistical perspective on machine learning}).
Upon looking into statistics and machine learning textbooks one often finds the
assumption of independent and identically distributed (i.i.d.) data points 
as the expression of randomness \citep[p. 207]{casella2002statistical} \citep[p. 4]{devroye2013probabilistic}.

We adopt Von Mises' differing account of randomness.
The expression
of randomness \emph{relative} to the problem at hand, particularly in settings with
data models such as statistics, turns out to be substantial.

\subsection{Orthogonal Perspectives on Randomness as Independence in Machine Learning and Statistics}
\label{orthogonal perspectivea on indepencence-randomness in machine learning and statistics}

Von Mises defined a random sequence as a sequence which is
statistically independent to a (pre-specified) set of selection rules 
respectively other
sequences. In contrast, an i.i.d.-sequence consists of elements each 
statistically independent to all others. 

Both definitions are stated in terms of statistical independence. But, the relationship
of independence and randomness in terms of i.i.d.~and in 
Von Mises' theory differ substantially.
Von Mises' randomness is stated relative with respect to a set of selection
rules. Furthermore, it is stated between sequences, respectively collectives.
Whereas, in an i.i.d.~sequence randomness is expressed between random variables.
The randomness definitions are in an abstract sense ``orthogonal.'' We consider a concrete
example for better understanding.
\begin{description}
    \item \textbf{Horizontal Randomness.} Let $\Omega = \mathbb{N}$ be a penguin colony. Let $s,
f$ be two attributes of a penguin, namely sex and whether a penguin has the
penguin flu or not. Mathematically: $s\colon \Omega \rightarrow \{ 0,1\}$, $f\colon
\Omega \rightarrow \{ 0,1\}$. So, penguins are individuals $n \in \Omega$ which we do not know
individually, but we know some attributes of them. Suppose we are given a
sequence $f(1), f(2), f(3), \ldots $ of flu values with
existing frequency limit.
This allows us to state randomness of $f$ with
respect to the corresponding sequence of sex values $s$,
containing infinitely many ones and having a frequency limit, by:
the sequence of sex values $s$ is
admissible on $f$. Respectively, $s$
and $f$ are statistically independent of each other.
In the context of colony $\Omega$ a
penguin having flu is random with respect to the sex of the penguin.

    \item \textbf{Vertical Randomness.} This is different to the i.i.d.-setting in which
each penguin $i \in \mathbb{N}$ obtains its own random variable $F_i\colon \Omega \rightarrow \{ 0,1\}$
on some probability space
$(\Omega, \mathcal{F}, P)$. Here, $F_i$ encodes whether penguin $i$ has the penguin flu or not.
The sequence $F_1, F_2, F_3, \ldots$ somehow represents the colony.
The included random variables share their distribution
and are statistically
independent to each other. The attribute flu is not random with respect to
the attribute sex
here, but the penguins are random with respect to
each other. The random
variables are (often implicitly) defined on a standard
probability space on $\Omega$. The set $\Omega$ here does \emph{not} model the colony.
It shrivels to an abstract source of
randomness and probability.

\end{description}
The choice of perspective, horizontal or vertical, on randomness expressed
as statistical independence is a question of the data model.
The two types of randomness definitions are distinct in a number of ways.
For a summary see Table~\ref{tab:horizontal-vertical-randomness}. 
Most importantly, horizontal randomness is inherently expressed
\emph{with respect to} some mathematical object.
Vertical randomness lacks this explicit relativization.
This typification of horizontal and vertical, mathematical definitions of randomness is actually
more broadly applicable.
\begin{table}[ht]
    \centering
    {\small 
    \begin{tabular}{p{7cm}ll}
        & \textbf{Horizontal Randomness} &  \textbf{Vertical Randomness}\\ \hline
        Data points are modelled as: &
        Evaluations of RVs & RVs
        \\[1mm] %\rowcolor[HTML]{D0D0D0} 
        Mathematical definition of randomness of:  & Sequences & Sequences of RVs\\[1mm]
        Explicit relativization: & Yes & No
    \end{tabular}
    \caption{Typification of horizontal and vertical randomness (``RV''=``random variable'').}
    \label{tab:horizontal-vertical-randomness}
    }
\end{table}

To the set of vertical randomness notions one can add:
exchangeability \citep{de1929funzione}, $\alpha$-mixing, $\beta$-mixing \citep{steinwart2009learning}
and possibly many more. The set of horizontal randomness notions is spanned
up by an entire branch of computer science and mathematics: algorithmic randomness.

Algorithmic randomness poses the question whether a sequence is random or not.
This question arose in \citep{mises1919grundlagen} within the
attempt to axiomatize probability theory \citep[p. 3]{bienvenu2009history}.
In algorithmic randomness further definitions
of random sequences have been proposed. For the sake of simplicity the considered
sequences consist only of zeros and ones.

Four intuitions for random sequences crystallized \citep[pp.280ff]{porter2012mathematical}:
typicality, incompressibility, unpredictability and independence (see Appendix~\ref{sec:4-int-randomness}).
For our purposes, the key point to note is that a random sequence is 
typical, incompressible, unpredictable or independent
with respect to ``something'' (they are all relativised in some way).
Each of these intuitions has been expressed in various mathematical terms.
In particular, formalizations of the same intuitions are not 
necessarily equivalent, and formalizations
of different intuitions sometimes coincide or are logically related 
(see Appendix~ \ref{sec:rel-diff-rand}).\footnote{For an
overview of algorithmic randomness see 
\citep{uspenskii1990can, muchnik1998mathematical, volchan2002random, downey2010algorithmic}.}
We mainly stick to the intuition of independence in this paper.
A random sequence is independent of ``some'' other sequences \citep{mises1919grundlagen, church1940concept}.

%%%%%%%%%%%%%%%%%%%%%%%%%%%%%%%%%%%%%%%%%%%%%%%%%%%%%
\subsection{Relative Randomness Instead of Absolute, Universal Randomness}
\label{relative randomness instead of absolute, universal randomness}
The definition of randomness for sequences
is inherently  relative. Even though,
the notion is relative with respect to ``something,''
most of the effort has been spent on
finding \emph{the} set of statistically independent 
sequences defining randomness
\citep{church1940concept, muchnik1998mathematical}.\footnote{The analogous
observation holds for all four intuitions (see Appendix~\ref{sec:4-int-randomness}
and \citep{muchnik1998mathematical}).}

Naively, one could attempt to define a random sequence as: a
sequence is random if and only if it is independent with respect to \emph{all} sequences.
However, this approach is doomed to fail. There is no sequence fulfilling
this condition except for trivial ones such as endless 
repetitions of zeros or ones (see Kamke's critique of von
Mises' notion of randomness \citep{lambalgen1987mises}).

So instead,
research focused on computability expressed in
various ways (because it was felt by those investigating these matters that 
computability was somehow given, or more primitive, and thus a natural way
to resolve the relativity of the notion of randomness). 
Intuitively, randomness is considered the antithesis of computability
\citep[p. 288]{porter2012mathematical}: something which is
computable is not random. Something which is random is not computable.
If we then informally update the definition above we obtain:
a sequence is random if and only if it is
independent with respect to all computable sequences \citep{church1940concept}\footnote{To
be precise, a random sequence following
\citet{church1940concept} is independent to all \emph{partially computable} selection rules following
Von Mises (see Footnote~\ref{footnote:vonmises-selectionrules}).}.
Analogous to \emph{the} definition of computability
\citep[p. 165]{porter2012mathematical}, this is taken as an argument 
for the existence of \emph{the} definition of randomness \citep[p. 287]{porter2012mathematical}.

In our work, we argue towards a relativized conception of randomness in line
with work by \cite{porter2012mathematical}, \cite{humphreys1977randomness} 
and \cite{mises1964mathematical}\footnote{
    \citet{humphreys1977randomness} presented randomness as relativized to a probabilistic hypothesis
    or reference class. \citet[p. 169]{porter2012mathematical}
    even postulated the ``No-Thesis''-Thesis: Any notion of randomness neither defines a
    well-defined collection of random sequences nor captures all mathematical
    conceptions of randomness; confer the logic of ``essentially contested'' 
    concepts \citep{gallie1955essentially}, which, presumably, are unavoidably 
    contested for the same reason.}.
A \emph{relative} definition of randomness is a definition of randomness
which is \emph{relative} with respect to the problem under consideration.\footnote{In machine learning,
a problem under consideration is, for instance, animal classification via neural networks.}
In contrast, an \emph{absolute and universal} definition of randomness would preserve its
validity in all problems. It presupposes the existence of \emph{the} randomness.

\emph{Relative} randomness with respect to the problem which we want to describe
aligns to Von Mises' theory of probability and randomness.
Von Mises emphasized the \emph{ex ante}
choice of randomness \citep[p. 89]{mises1981probability} \emph{relative} 
to the problem at hand \citep[p. 12]{mises1964mathematical}.
First, one formalizes randomness with respect to the underlying problem,
then one can consider a sequence to be random or not.
Otherwise, if we are given a sequence, it is easy
to construct a set of selection rules, such that the sequence is random
with respect to this set.\footnote{This idea is transferable to other
intuitions of randomness, for instance \citep{ville1939etude}.} This, however, undermines the
concept of randomness, which should capture the pre-existing typicality, incompressibility, unpredictability 
or independence of a sequence (cf. \citep[p. 321]{vovk1993logic}).
Von Mises' randomness intrinsically possesses a modeling character, similar to our needs in
machine learning and statistics.

Given its role as modeling assumption in statistics, randomness lacks
substantial justification to be expressed in any \emph{absolute and universal} manner 
in this context. 
Neither,
there are reasons why computability\footnote{Computability is often taken (at 
least by computer scientists) as a purely mathematical notion, 
detached from the world. An alternate view, close in spirit 
to Von Mises, is that computation is part of physics, and thus
needs to be viewed in a \emph{scientific}, and not merely 
\emph{mathematical} manner \citep{deutsch2011beginning}.
} is the only mathematical,
expressive way to encode one of the four intuitions of randomness.
The i.i.d.~assumption, an \emph{absolute and universal}
definition of randomness, does not fits thpi purpose.
To appropriately model data we require adjustable notions of randomness. 
Otherwise, we restrict our modeling choice without reason
or gain.\footnote{This is not entirely true, as specific computability
notions of randomness and the i.i.d.-assumption deliver convergence results
for random sequences, which can be used to guarantee low estimation
error in the long run.}

Equipped with the interpretation of 
statistical independence as
randomness we now return to our motivation for investigating
statistical independence. ML-friendly fairness criteria are built upon
statistical independence. In contrast to Kolmogorov,
Von Mises' statistical independence
transparently refers to a concept of independence in the real world.
To clarify the meaning of fairness expressed as statistical independence,
we directly apply Von Mises' independence to the fairness criteria 
listed in Section~\ref{fairness as statistical indepedencen} in the following.

\section{Von Mises' Fairness}
\label{von mises fairness}
With Von Mises' definition of statistical independence we have a notion
at our disposal which is conceptually focused on a more ``scientific'' 
perspective  (i.e.~making claims about the world) of
statistical concepts. Since it is mathematically related
to Kolmogorov's standard account of statistical
independence, Kolmogorov's definition can, at many
places, be easily replaced by Von Mises' definition.

Let us denote the three presented
fairness criteria in a
Von Mises' way (cf. Section~\ref{Kolmogorov's Independence versus Von Mises' Independence}).
\begin{definition}[Fairness as Statistical Independence]
\label{def:fairness as statistical independence}
    A collective $x\colon\mathbb{N}\rightarrow \{ 0,1\}$ (with respect to  a 
    family of selection rules $\mathcal{S}$) is \emph{fair
    with respect to a set of sensitive groups} $\mathcal{G}=\{s^j\colon\mathbb{N}\rightarrow\{0,1\} |  j\in J\}$
    if
    \begin{align*}
        x  \perp s^j \ \ \forall j\in J
    \end{align*}
\end{definition}
The 0-1-sequences $s^j$ determine for each individual $i$ whether it is part of
the group or not (according to whether $s^j(i)=1$ or $s^j(i)=0$). 
We call these groups ``sensitive,'' as these are the
groups which are of moral and ethical concern. In philosophical literature
these groups are often called ``socially salient groups'' 
\citep{Altman2020discrimination, loi2019philosophical}.\footnote{
Intersections of sensitive groups are not necessarily independent.}
We see that the connection between Von Misean independence and fairness arises from the 
observation that the set of sensitive groups  $\mathcal{G}$
\emph{is} a family of selection rules, so that if $\mathcal{G}\subseteq\mathcal{S}$, then indeed the collective 
$x$ will be fair for $\mathcal{G}$.

Following Von Mises' interpretation of independence, the given definition reads
as follows: we assume, we are in the idealized setting of infinitely many individuals with
values $x_i$, e.g. binary predictions. The predictions are fair if and only if there is no
difference in counting
the frequency of $1$-predictions in the entirety or in a sensitive group.
(For an illustration see Appendix~\ref{example fair penguin colony}.)
A proper conceptualization of fairness requires such immediate semantics, but
a purely mathematical theory of probability cannot offer these
(see Section~\ref{statistical independence as we know lacks semantics}).

Each of the three fairness criteria is captured in 
Definition~\ref{def:fairness as statistical independence};
the choice of fairness criterion manifests in the collective under consideration:
\begin{description}
    \item[Independence] The collective $x\colon\mathbb{N}\rightarrow \{ 0,1\}$ consists 
         of predictions; i.e. $\{ 0,1\}$ is the set of predictions. 
    \item[Separation] The collective $x\colon\mathbb{N}\rightarrow \{ 0,1\}$ is obtained via 
          the subselection of predictions  based on the sequence of true labels corresponding to the 
             predictions.\footnote{``Selecting with respect to'' is ``conditioning on.''}
    \item[Sufficiency] The true labels are subselected by predictions.
\end{description}

The three fairness criteria Independence, Separation and Sufficiency encompass
a large part of fair machine learning \citep[p. 45]{barocas2017fairness}.
Von Mises' statistical independence gives a consistent interpretation
to all of them.
In fact, Von Mises' independence opens the door to further
investigations.
To this end, we we recapitulate the strong linkage between
statistical independence and randomness in Von Mises' theory.

\section{The Ethical Implications of Modeling Assumptions}
\label{the ethical implications of modeling assumptions}
Machine learning methods try to model data in complex ways.
Derived statements, such as predictions, then potentially get applied
in society. 
In these cases one is obliged to ask which ought-state the
machine learning model
should reflect \citep{passi2019problem, raz2021group}.
To enable a justified choice,
statistical concepts in machine learning require relations
to the real world. Furthermore, modeling
even requires understanding of the entanglement of societal
and statistical concepts.

We proposed one specific \emph{meaningful} definition of
statistical independence which can be directly applied 
to the three observational fairness criteria from fair machine
learning. In addition, this Von Mises' independence
is key to a relativized notion of randomness.
Pulling these threads together, we are now able to
establish the following link: \emph{Randomness is fairness. Fairness is randomness.}

\subsection{Randomness is Fairness. Fairness is Randomness.}
The concepts fairness and randomness frequently appear jointly: \cite{Broome1984} 
argues that a 
 random allocation of goods is fair under certain conditions.
 Literature on sortition argues for just representation of society by
random selection of people 
\citep{parker2011randomness,stone2011luck}.\footnote{Representativity here 
can be interpreted as typicality,
thus one of the four intuitions for randomness.}
\citet[p. 633]{benett2011defining} even states that randomness encompasses fairness.

With Von Mises' axiom~\ref{def:randomness a la Von Mises} and Definition~\ref{def:fairness as statistical independence}
we can now tighten the conceptual
relationship of fairness and randomness. The proposition directly follows
from the definition of randomness respectively
fairness in the sense of Von Mises.
\begin{proposition}[Randomness is fairness. Fairness is randomness.]
\label{prop:fairness is randomness}
    Let $x$ be a collective with respect to
    $\emptyset$ (the empty set). It is fair with respect to a set of sensitive groups
    (0-1-sequences) $\{ s^j \}_{j \in J}$, if and only if it is
    is random with respect to  $\{ s^j \}_{j \in J}$.
\end{proposition}
The given proposition establishes a helpful link. It gives insights into
both of the concepts. In particular, it substantiates the relativized conception
of randomness in machine learning as it presents randomness as an ethical choice.

\subsubsection{Randomness as Ethical Choice}
\label{randomness as ethical choice}
Randomness in machine learning is a modeling assumption (Section~\ref{a statistical perspective on machine learning}).
Fairness is an ethical choice.\footnote{More specifically, the choice of operationalized
fairness, one of the fairness criteria, and the choice of groups.}
In light of Proposition~\ref{prop:fairness is randomness} randomness gets
an ethical choice and fairness a modeling assumption.
We now further detail this perspective.

We assume that we are given a fixed set of selection rules, which defines
``the'' randomness.
As far-fetched as this may sound, if we, for example, accept the so called
Martin-Löf randomness
as \emph{absolute and universal} definition, then we exactly do this and 
fix the set of selection rules to the partial computable ones
(see Appendix~\ref{sec:proto-absuniversal-rand}).
A sequence which is random with respect to this specified set of selection rules
is fair with respect to the groups defined by the selection rules.
Rephrased in terms of Martin-Löf randomness: a Martin-Löf random sequence
is fair with respect to all partial computable groups.
Only non-partial-computable groups (respectively sequences) can be
discriminated against in this setting.
If we interpret statistical independence as fairness (Section~\ref{fairness as statistical indepedencen}),
then fairness is as \emph{absolute and universal} as randomness here.
Where did the ``essentially contested'' nature of fairness \citep{gallie1955essentially} leave the picture?

The set of admissible selection rules
specifies the choice of sensitive groups, which indeed is a fraught and contestable 
choice \cite[Section H.3]{menon2018cost}.
Thus each selection rule gets ethically loaded. 
Furthermore, the choice of collective, which we consider as random, fixes the fairness criterion.
In summary, the determination of randomness is analogous to the determination
of fairness.

However one defines randomness,
it is an ethical choice. For symmetry 
reasons one can equivalently state in machine learning: 
fairness is a modeling assumption.
The randomness assumption has an ethical, moral and potentially
legal implication. We need non-mathematical, contextual
arguments to each problem at hand which justify the adjustable and
explicit randomness assumptions.

Given that randomness is an ethical choice, an absolute, universal conception of
randomness counteracts any ethical debate in machine learning.
Discussions about sexism,
racism and other kinds of discrimination and injustice persist over time
without ever arrogating the discovery of ``the'' fairness \citep{gallie1955essentially}.
But if ``the'' randomness as statistical independence would exist, 
then ``the'' fairness as statistical independence
would be an accessible notion. 
For illustration, we reconsider Martin-Löf randomness. A Martin-Löf random sequence
is independent, respectively fair, to the set of all partial computable selection rules.
But, it is completely unclear what the ethical meaning of partial computable groups is.
And, it remains unsolved whether the groups given by gender are partial computable,
when we desire to be fair with respect to them.
We conceive Proposition~\ref{prop:fairness is randomness} as further counterargument
to an absolute, universal definition of randomness.
Randomness is, like fairness, better interpreted as a \emph{relative} notion.

Further concluding,
the equivalence of randomness and fairness highlights 
the deficiency of fairness notions in machine learning.
The equivalence only holds due to the very
reductionist perspective on fairness
in fair machine learning.
Despite their regular co-occurrence \citep{Broome1984, parker2011randomness} 
\citep[p. 633]{benett2011defining}, fairness and randomness
are more multi-facetted and non-overlapping concepts
as illustrated in Proposition~\ref{prop:fairness is randomness}.

\subsubsection{Fairy Tales of Fairness: ``Perfectly Fair'' Data}
With the relationship between fairness and randomness in mind, we now turn towards
random data as primitive.
Discussions in fair machine learning sometimes 
seemingly presume the existence
of ``perfectly fair'' data (e.g. as highlighted in \citep[p.~134]{raz2021group}),
as if fair machine learning merely tackles
the cases where ``perfectly fair'' data is not available.

We interpret ``perfectly fair'' data as a collective with respect to all possible
selection rules. The data does not depend on any (sensitive) group at all. 
In other words, ``perfectly fair'' data is
``totally random'' data. As we saw in Section~\ref{relative randomness instead of absolute, universal randomness}
this is self-contradicting except
of the trivial constant case. ``Perfectly fair'' data does not exist or is
statistically useless.

\subsection{Demanding Fairness is Randomization: Fair Predictors are Randomizers}
In practice, it is often unreasonable to \emph{assume} random or fair data as in
Proposition~\ref{prop:fairness is randomness}. Instead one \emph{demands} for fairness respectively
randomness of predictions.  In these settings,
 fair machine
learning techniques are deployed to exhibit ex post fulfillment of fairness criteria.

We assume for the following discussion that the collective $x$
consists of predictions, as in the fairness criteria Independence or Separation.
Fair machine learning techniques enforce statistical
independence of
predictions and sensitive attributes. Rephrased, fair machine learning
techniques actually introduce randomness post-hoc into the predictions.
Thus, fair machine learning techniques can potentially be 
interpreted as randomization
techniques.

\subsubsection{Fairness-Accuracy Trade-Off --- Another Perspective}
We noticed that fair predictions are random predictions with respect
to the sensitive attribute. In contrast, accurate predictions exploit all dependencies 
between given attributes and predictive goal, including the sensitive attributes.
Thus, in fair machine learning morally wrongful discriminative potential of sensitive
attributes is thrown away by purpose. On these grounds, it is not surprising that 
an increase in fairness respectively randomness (usually) goes hand in hand with a decrease
in accuracy \citep{wick2020unlocking}. Randomization of predictions leads to the
so called fairness-accuracy trade-off.
\\

Concluding, via Von Mises' axiomatization we established: 
\emph{Randomness is fairness. Fairness is randomness}.
Exploiting this new perspective, we unlock another perspective on fair predictors as randomizers,
demonstrate the nonexistence of ``perfectly fair'' data
and treat randomness as an ethical choice, which
can be neither universal nor total. In particular, the ``essentially contested'' nature 
of fairness is tied to the ``essentially \emph{relative}'' nature of randomness.

\section{Conclusion}

Fair machine learning attained an increasing interest in the last years.
However, its conceptual maturity lags behind. In particular, the 
interplay between data, its mathematical representation and their relation
to fairness is encompassed by a veil of nescience. In this paper, we 
contribute towards a better understanding of randomness and fairness in
machine learning.

We started from the most commonly used definition of statistical
independence and questioned
its representation due to a lack of semantics.
Generally, we observe that in machine learning, as in statistics, probability
and its related concepts should be interpreted as modeling assumptions
about the world (of data). Von Mises aimed for exactly this ``scientific''
perspective on probability theory.
We lean on his statistical independence, which clarifies the relation
to the real world, and his definition of randomness, which is \emph{relative}
and orthogonal to the i.i.d. assumption, but similarly expressed as
statistical independence.
Then by the three fairness criteria in machine learning we obtain
a further interpretation of independence, which we finally exploit to argue
for a \emph{relative} conception of randomness, randomness as an ethical choice in
machine learning and fair predictors as randomizers. 
% \emph{Randomness is Fairness. Fairness is randomness} is a
% fruitful perspective with: probably more implications waiting to be find out.

\subsection{Future Work: Approximate Randomness and Fairness, Randomness as Fairness via Calibration}
Despite future conclusions in-between the topics fairness and
randomness in other research subjects as machine learning, we claim that
a significant dimension is missing in the present discussion. Practitioners usually deal 
with approximate versions of randomness, statistical independence
or fairness. Yet, approximation spans another dimension of choice
beset with pitfalls \citep{putnam1990meaning, lohaus2020too}.
Several questions ranging from the choice of approximation to
the interference of concepts arise. Future work should detail the
implications of this choice.

Second, we conjecture that ``\emph{Randomness is Fairness. Fairness is randomness.}'' can be 
substantiated via the intuition of unpredictability. Starting from 
\citep{shafer2019game} definition of unpredictability randomness, which is
closely related to the calibration idea presented in \citep{dawid2017individual},
we can bridge to fairness as calibration as given in \citep{Chouldechova2017}. In
particular, this chain of arguments entails a more thorough discussion
of the concepts of individual versus group fairness in machine learning \citep{binns2020apparent}.
As a subproblem, which is contained therein,
the categorization into (sensitive) groups in fair machine learning deserves its own work.

Third, regarding a more thorough definition of statistical independence within the 
fairness criteria, we are convinced
that a subjectivist interpretation of probability might reveal yet
another perspective on the problem. We assume that the interplay between different interpretations
of probability and ethical concepts such as fairness still leaves room for many 
important investigations.

Last but not least, we already referred to sortition literature
and random allocation. The somewhat different relation between fairness and randomness
in this literature leads us to speculate that further
fruitful discussions between the two concepts may develop.
\\

In the jungle of statistical concepts such as probability, uncertainty, 
randomness, independence etc. further relations to social and ethical concepts wait to
be brought to light. And machine learning research should care:
\begin{quote}
The arguments that justify inference from a sample to a population should
explicitly refer to the variety of non-mathematical considerations
involved.\\
\citep[p.~11]{battersby2003rhetoric}
\end{quote}

\section*{Acknowledgements}
Many thanks to Christian Fröhlich, Eric Raidl, Sebastian Zezulka, Thomas Grote
and Benedikt Höltgen
for helpful discussions and feedback.
Futhermore, the authors thank all participants of the
Philosophy of Science meets Machine Learning Conference 2022 in Tübingen,
for all helpful comments and debates.

This work was was funded in part by the Deutsche Forschungsgemeinschaft
(DFG, German Research Foundation) under Germany’s Excellence Strategy –-
EXC number 2064/1 –- Project number 390727645; it was also supported by the 
German Federal Ministry of Education and Research (BMBF): Tübingen AI Center.
The authors thank the International Max Planck Research School for
Intelligent System (IMPRS-IS) for supporting Rabanus Derr. 

\bibliographystyle{plainnat}
\bibliography{library}

\appendix
\section{Generalized Von Misesean Probability Theory}
\label{generalized Von Misesean probability theory}
In this appendix we outline a theory of probability subsuming that of Kolmogorov
and Von Mises.
\subsection{Kolmogorov's notion of Independence}
Kolmogorov axiomatized probability theory in his book \citep{kolmogorov2018foundations}
in a measure theoretical way. He defined a
probability space $(\Omega, \mathcal{F}, P)$ as a measure space with base
set $\Omega$, $\sigma$-algebra $\mathcal{F}$ and normalized measure $P$.
Events are elements of $\mathcal{F}$, i.e. subsets of $\Omega$, which
obtain a probability via $P$. Statistical independence is defined as a
specific assignment of probability to an intersection event.
\begin{definition}[Kolmogorov's Definition of Statistical Independence of Events  \protect{\citep[p. 9 Def. 1]{kolmogorov2018foundations}}]
\label{def:kolomogorov independence}
Let $(\Omega,
	\mathcal{F}, P)$ be a probability space. Two events $A, B \in
	\mathcal{F}$ are called statistically independent iff
	\begin{align*}
	    P(A \cap B) = P(A) P(B).
	\end{align*}
\end{definition}
As we highlighted above, Kolmogorov's axiomatization is, despite its success,
not the only mathematical theory of probability. Specifically,
one can weaken the structure of the probability space and still
work with concepts such as statistical independence, expectation, conditioning (e.g. \citep{gudder1969quantum, chichilnisky2010foundations, narens2016introduction}).

\subsection{Finitely Additive Probability Space}
\label{finitely additive probability space}
We introduce a weaker measure structure, which we call finitely additive probability space.
Interestingly, this weaker structure includes the axiomatization of
Kolmogorov and Von Mises as special cases.
We define a finitely additive probability space modified from 
\protect{\citep[Def. 2.1.1 (7)]{rao1983theory}}) as
\begin{definition}[Finitely Additive Probability Space]
\label{def:finaddprobspace}
The tuple $(N,
	\mathcal{A}, \nu)$ is called a \emph{finitely additive probability
	space} for a base set $N$, a set of measurable sets $\mathcal{A}
	\subset \mathcal{P}(N)$ containing the empty set $\emptyset \in
	\mathcal{A}$ and a finitely additive probability measure $\nu:
	\mathcal{A} \rightarrow [0,1]$ satisfying the following
	conditions:\\ (1) $\nu(\emptyset) = 0$,\\ (2) if $A_1, A_2, A_1
	\cup A_2 \in \mathcal{A}$ and $A_1 \cap A_2 = \emptyset$ then
	$\nu(A_1 \cup A_2) = \nu(A_1) + \nu(A_2)$.
\end{definition}
Observe that this definition does not impose any structural restrictions on the
set of subsets $\mathcal{A}$.

Kolmogorov's probability space is certainly a specific finitely additive probability
space in our sense, as every set $\sigma$-algebra contains the empty set and every
countably additive probability is finitely additive.

Analogously, Von Mises implicitly uses a finitely
additive probability space. This space is given by $(\mathbb{N},
\mathcal{A}_{\vM}, \nu_{\vM})$, where $\mathbb{N}$ are the natural numbers
and $\mathcal{A}_{\vM}, \nu_{\vM}$ are defined in the following.

First, we consider the finitely additive base measure
\begin{align*}
    \nu_{\vM}(A) := \lim_{n \rightarrow \infty} \frac{|A \cap \mathbb{N}_n|}{n},
\end{align*}
where $A \subset \mathbb{N}$ and $\mathbb{N}_n = \{ 1, ..., n\}$.
$\nu_{\vM}$ is called the ``natural density'' in the number theory literature \citep[p. 256]{nathanson2008elementary}.
From this definition
is not clear whether the given limit exists. Thus, we define the set of
measurable sets by
\begin{align*}
    \mathcal{A}_{\vM} := \{A: \nu_{\vM}(A) \text{ exists} \},
\end{align*}
which is called the ``density logic'' in \citep{ptak2000concrete}. It is a
pre-Dynkin-system \citep{schurz2008finitistic}.

We generalize Kolmogorov's definition of statistical independence of events to
finitely additive probability spaces.
\begin{definition}[Statistical Independence of Events on a Finitely Additive Probability Space]
Let $(N, \mathcal{A}, \nu)$ be a finitely additive probability space. Two
measurable sets $A, B \in \mathcal{A}$ are independent iff
\begin{enumerate}
    \item $A \cap B \in \mathcal{A}$
    \item $\nu(A \cap B) = \nu(A) \nu(B)$
\end{enumerate}
\end{definition}
Observe that the first condition is naturally fulfilled in Kolmogorov's
$\sigma$-algebra. In the case of Von Mises' density logic, this constraint
is strict. A pre-Dynkin-system is not closed under arbitrary intersections.

\subsection{Von Mises' Admissibility and Kolmogorov's Independence are Analogues}
\label{Von Mises independence and kolmogorovs independence are analogues}
We now reconsider the definition of collectives and selection rules. Both are
$0,1$-sequences on the natural numbers, with the restriction that selection
rules contain infinitely many $1$'s and their frequency limit does not have
to exist. Both sequences can be interpreted as indicator functions on the
natural numbers. So for  $x \colon \mathbb{N} \rightarrow \{ 0,1\}$ and  $s\colon \mathbb{N} \rightarrow \{ 0,1\}$
we write $X=x^{-1}(1)$ and $S=s^{-1}(1)$ for the corresponding subsets of the natural numbers.

We want to show that Von Mises' admissibility condition is equivalent to
the given definition of statistical independence on finitely additive
probability spaces. Actually, the equivalence only holds for the slightly
restricted case in which the selection rule itself possesses converging frequencies.

\begin{theorem}[Admissibility in Von Mises setting implies Statistical Independence
on Finitely Additive Probability Spaces]
\label{thm:admissability gives stat independence}
Let $x$ be a
collective with respect to $s$. Suppose furthermore that
$s$ has a converging frequency limit.
Then $X$ and
$S$, the indexed sets corresponding to the collective 
(respectively selection rule)
on the finitely additive probability space
$(\mathbb{N}, \mathcal{A}_{\vM}, \nu_{\vM})$  are statistically independent.
\end{theorem}
\begin{proof}
We know that
\begin{align*}
    \lim_{n \rightarrow \infty} \frac{|\{i : x(i) = 1 \text{ and } s(i) = 1,
    1 \le i \le n\}|}{|\{j : s(j) = 1, 1 \le j \le n\}|} = \lim_{n
    \rightarrow \infty} \frac{|\{i : x(i) = 1, 1 \le i \le n\}|}{n}
\end{align*}
which we can rewrite as
\begin{align*}
    \lim_{n \rightarrow \infty} \frac{|X \cap S \cap \mathbb{N}_n|}{|S \cap
	    \mathbb{N}_n|} = \lim_{n \rightarrow \infty} \frac{|X \cap
		    \mathbb{N}_n|}{n}.
\end{align*}
This gives 
\begin{align*}
    \nu_{\vM}(X \cap S) &= \lim_{n \rightarrow \infty} \frac{|X \cap S \cap
	    \mathbb{N}_n|}{n}\\
    &= \lim_{n \rightarrow \infty} \frac{|X \cap S \cap \mathbb{N}_n|}{|S
	    \cap \mathbb{N}_n|}\frac{| S \cap \mathbb{N}_n|}{n}\\
    &= \lim_{n \rightarrow \infty} \frac{|X \cap S \cap \mathbb{N}_n|}{|S
	    \cap \mathbb{N}_n|} \lim_{n \rightarrow \infty}  \frac{| S \cap
		    \mathbb{N}_n|}{n}\\
    &= \lim_{n \rightarrow \infty} \frac{|X \cap \mathbb{N}_n|}{n} \lim_{n
    \rightarrow \infty} \frac{| S \cap \mathbb{N}_n|}{n}\\
    &= \nu_{\vM}(X) \nu_{\vM}(S),
\end{align*}
by help of a standard result for the multiplication of sequence limits \citep[Theorem 3.1.7]{liskevich2014analysisI}.
\end{proof}

\begin{theorem}
\label{thm:independent events give admissible selection rules}
Let $X$ and $S$ be two statistically independent events on the finitely additive
probability space $(\mathbb{N}, \mathcal{A}_{\vM}, \nu_{\vM})$ with
$\nu_{\vM}(S) > 0$, then the corresponding collective $x$, indicator function of $X$,
has the admissible selection rule $s$, indicator function of $S$.
\end{theorem}
\begin{proof}
It is given that 
\begin{align*}
    \lim_{n \rightarrow \infty} \frac{|X \cap S \cap \mathbb{N}_n|}{n} =
    \nu_{\vM}(X \cap S) = \nu_{\vM}(X) \nu_{\vM}(S) = \lim_{n \rightarrow
    \infty} \frac{|X \cap \mathbb{N}_n|}{n} \lim_{n \rightarrow \infty}
    \frac{| S \cap \mathbb{N}_n|}{n}.
\end{align*}
Furthermore $\nu_{\vM}(S) > 0$ implies that the corresponding selection
rules selects infinitely many elements and $\lim_{n \rightarrow \infty}
\frac{n}{|S \cap \mathbb{N}_n|} = \frac{1}{\nu_{\vM}(S)}$.

This implies
\begin{align*}
    \lim_{n \rightarrow \infty} \frac{|X \cap S \cap \mathbb{N}_n|}{|S \cap
	    \mathbb{N}_n|} &= \lim_{n \rightarrow \infty} \frac{|X \cap S
	    \cap \mathbb{N}_n|}{n} \frac{n}{|S \cap \mathbb{N}_n|}\\
    &= \lim_{n \rightarrow \infty} \frac{|X \cap S \cap \mathbb{N}_n|}{n}
	    \lim_{n \rightarrow \infty} \frac{n}{|S \cap \mathbb{N}_n|}\\
    &= \lim_{n \rightarrow \infty} \frac{|X \cap \mathbb{N}_n|}{n} \lim_{n
	    \rightarrow \infty} \frac{| S \cap \mathbb{N}_n|}{n} \lim_{n
	    \rightarrow \infty} \frac{n}{|S \cap \mathbb{N}_n|}\\ 
    &= \lim_{n
    \rightarrow \infty} \frac{|X \cap \mathbb{N}_n|}{n}.
\end{align*}
\end{proof}
 We note  some caveats regarding the preceding discussion.
\begin{enumerate}
\item We require the frequency limit for $s$
to exist in Theorem~\ref{thm:admissability gives stat independence}.
So the sequence $S$
is not an entirely general selection rule.

\item The condition $\nu_{\vM}(S) > 0$ in Theorem
\ref{thm:independent events give admissible selection rules} ensures
that we do not condition on measure zero events.
Furthermore, it guarantees at this point, that the indicator
function of $S$ is a selection rule containing infinitely many ones.
Besides that, its frequency limit exists.

\item Even though given here only for the binary case. The
argumentation can be extended to continuum labeled collectives
\citep[II.B]{mises1964mathematical}.

\item In our discussion, we focus on admissibility instead of statistical independence
in the sense of Von Mises, since our argument finally focuses
on the randomness interpretation of statistical independence.
Admissibility
and Von Mises' statistical independence are, despite the 
objects on which they are defined, equivalent. 

\item Statistical independence in Von Mises' setting and
Kolmogorov's setting are not equivalent. They are special cases of a
more general form of statistical independence. So they are mathematically
analogous under mild conditions. But we emphasize
the conceptual difference. Von Mises refused to define statistical
independence between mere events \citep[p. 35]{mises1964mathematical}.
Instead he demanded statistical independence to be defined between
collectives underlining that independence is a concept about aggregates,
the collectives, not single occurring events.
\end{enumerate}

\section{Four Intuitions of Randomness}
\label{sec:4-int-randomness}

\begin{description}
	\item[Typicality] A sequence is called random if it shares ``some''
		characteristics of any possible sequence. Martin-Löf and Schnorr
		formalized this idea via statistical tests \citep{martin1966definition, schnorr1977survey}.
	\item[Incompressibility] The information contained in a random sequences is
		(approximately) as large as the sequence itself. The
		sequences cannot be compressed by ``some'' procedure \citep{kolmogorov1965three, chaitin1966length, schnorr1972process, levin1973concept}.
	\item[Unpredictability] In a random sequence one can know all
		foregoing elements without being able to predict by ``some'' procedure the next element \citep{ville1939etude, muchnik1998mathematical, shafer2019game,frongillo2020memoryless, cooman2022randomness}. 
	\item[Independence] A random sequence is independent of ``some''
		other sequences \citep{mises1919grundlagen, church1940concept}.
\end{description}

\section{Relations between Randomness Definitions}
\label{sec:rel-diff-rand}
We briefly outline some of the known relationships between various notions of 
randomness.

% \subsection{Same Randomness Intuition - Differing Formalization}
%\label{Same Randomness Intuition - Differing Formalization}
On one hand, there are mathematical expressions
capturing the same randomness intuitions meanwhile being
mathematically distinct to each other.

For instance, the definition of a typical sequence following 
\citep{martin1966definition} implies
Schnorr's definition of typicality \citep{schnorr1971zufalligkeit, schnorr1977survey}
but not vice versa \citep[p. 143]{uspenskii1990can}.
Cooman and De Bock's imprecise unpredictability
randomness \citep{cooman2022randomness} is strictly more expressive than 
Vovk and Shafer's unpredictability randomness \cite[Section 1.1]{shafer2019game}
\citep[Theorem 37]{cooman2022randomness}.

% \subsection{Different Randomness Intuitions - Necessary, Sufficient or Equivalent Formalization}
% \label{Different Randomness Intuitions - Necessary, Sufficient or Equivalent Formalization}
On the other hand, there are mathematical expressions
capturing differing randomness intuitions meanwhile being
necessary, sufficient or even equivalent to each other.
For instance, the Levin-Schnorr theorem, simultaneously proven
by Levin \citep{levin1973concept} and Schnorr 
\citep{schnorr1972process, schnorr1977survey},
established an equivalence of
typicality following Martin-Löf \citep{martin1966definition} and 
incompressibility
following Levin \citep{levin1973concept} and Schnorr 
\citep{schnorr1972process} based on the idea of \citep{kolmogorov1965three} 
(e.g. see in \citep[Theorem 5.3]{volchan2002random} and references therein). 
Cooman and de Bock expressed Schnorr \citep{schnorr1971zufalligkeit, schnorr1977survey}
and Martin-Löf randomness \citep{martin1966definition} in terms of an
unpredictability approach \citep{cooman2022randomness}.
Muchnik showed in \citep{muchnik1998mathematical} that
all incompressible sequences, again following Levin and Schnorr's
prefix-complexity approach  \citep{schnorr1972process, levin1973concept}
are unpredictable in his sense. 

In his thesis \cite{ville1939etude} attempted to generalize the idea of
``excluding a gambling strategy'' (see Section~\ref{Von Mises theory of probability and randomness in a nutshell}) via a
game-theoretic approach. He showed \cite[p. 76]{ville1939etude} that for any set of
admissible selection rules $\mathcal{S}$ he could construct a gambling
strategy, more exactly a capital process associated to a
gambling strategy, which captures the same randomness
definition. The opposite direction is, however, impossible \cite[p. 39]{ville1939etude}.
But \cite{ambos1996resource} introduced a weaker form of \cite{ville1939etude}'s
randomness as unpredictability which is equivalent to \cite{church1940concept}'s
randomness as independence \citep[Section 12.3]{downey2010algorithmic}.
Finally, van Lambalgen observed an
abstract interpretation of ``random with respect to something'' as
``independent to'' in \citep{lambalgen1990axiomatization}.

\subsection{A prototypical, absolute and universal Notion of Randomness}
\label{sec:proto-absuniversal-rand}
One often referred \emph{absolute and universal} definition of randomness
is Martin-Löf's typicality approach \citep{buss2013probabilistic}.
Martin-Löf's randomness as typicality has been equivalently
formalized in terms of incompressibility and unpredictability
(see Section~\ref{sec:rel-diff-rand}). Furthermore, a sequence, which is Martin-Löf
random \citep{martin1966definition}, is statistically independent to all partial
computable selection
rules \citep{church1940concept} \citep[Theorem 11]{tadaki2014operational}.
Contrarily, not every collective
with respect to all partial computable selection rules is 
a Martin-Löf random sequence
\citep[p. 193]{blum2009einfuhrung}.
So Martin-Löf's definition is linked to all four intuitions.

\newpage
\section{Kolmogorov's versus Von Mises' Probability Theories in a Table}
%Table \ref{tab:kolmogorov-vs-vonmises-summary} summarizes the key differences between Kolmogorov's and Von Mises' probability theories. 

\begin{table}[h]
{\small
    \centering
    \begin{tabular}{p{2.2cm}|p{5cm}p{7.3cm}}
         & Kolmogorov & Von Mises \\ \hline
        fundamental structure & probability space $(\Omega, \mathcal{F}, P)$ & collectives $(x_i)_{i \in \mathbb{N}}$, implicitly $(\mathbb{N}, \mathcal{A}_{\vM}, \nu_{\vM})$\\[6mm]
        base set & (almost arbitrary) set $\Omega$ & $\mathbb{N}$\\[2mm]
        set of events & $\sigma$-algebra $\mathcal{F}$ & Dynkin-system $\mathcal{A}_{\vM}$\\[2mm]
        probability & finite positive measure & limit of frequency sequence\\[2mm]
        probability measure & countably additive & finitely additive $\nu_{\vM}$ \\[6mm]
        randomness & no explicit mathematical definition & explicit mathematical definition\\[2mm]
        statistical\ \ \ \ \ \ \ \ \ \  independence & factorization of joint distribution & frequency limit doesn't change under subselection\\[6mm]
        data model & each data point a random variable $X_i$ on  $(\Omega, \mathcal{F}, P)$ & each data point an element in a collective $(x_i)_{i \in \mathbb{N}}$
    \end{tabular}
    }
    \caption{Summary of main difference between Kolmogorov's and Von Mises' probability theories.}
    \label{tab:kolmogorov-vs-vonmises-summary}
\end{table}

\section{Penguin Colony Example for Fair Collective}
\label{example fair penguin colony}
\begin{figure}[ht]
\centering
\includegraphics[width=0.6\textwidth]{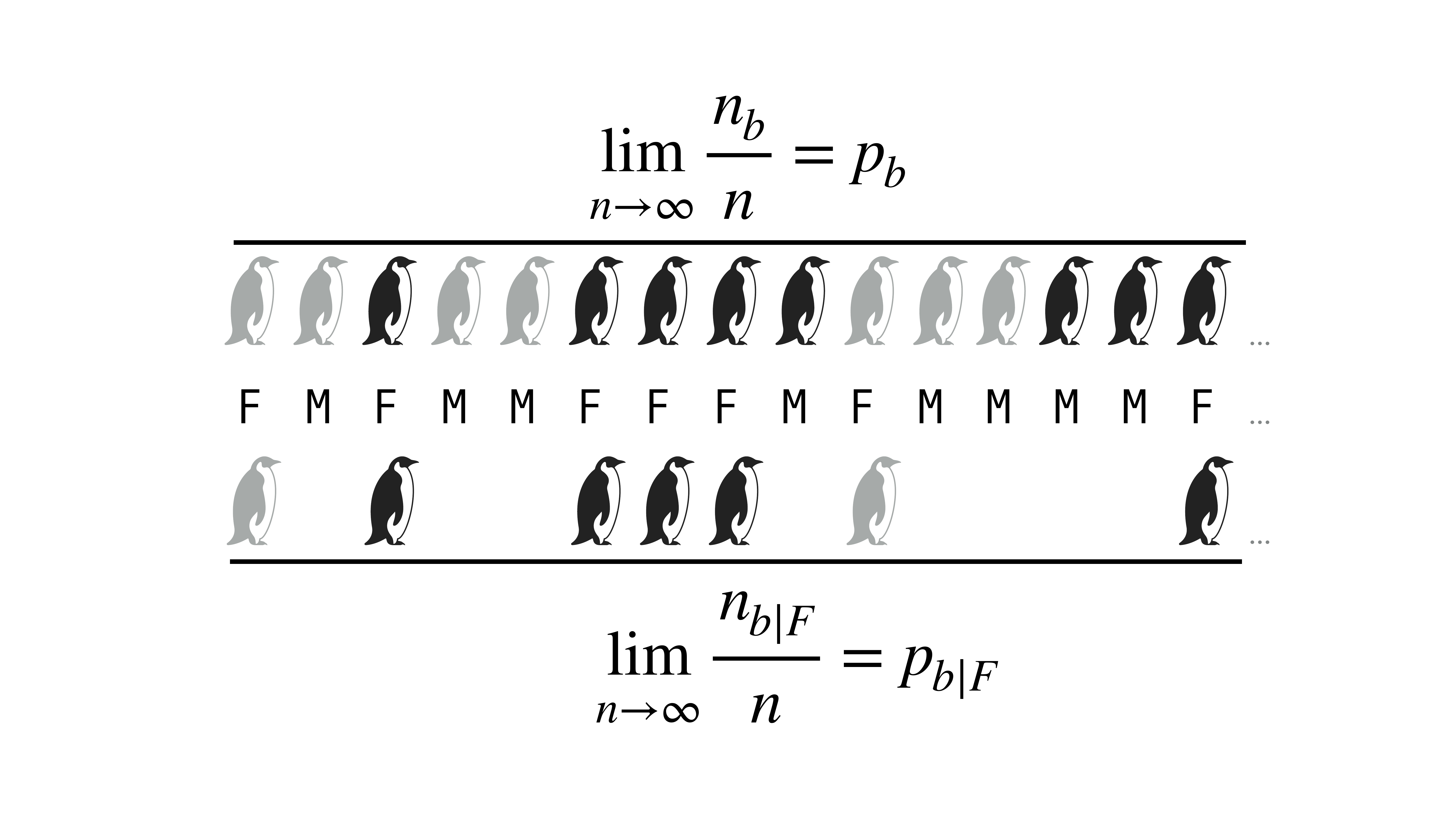}
\caption{Example for a subselection and fair collective. $n_b$ denotes the number of black penguins among the first $n$-penguins. Blackness of penguins
is distributed fairly with respect to sex if $p_b = p_{b |F}$.}
\end{figure}

\end{document}